\documentclass[preprint,12pt]{elsarticle}
\usepackage{epsfig}
\usepackage{amsmath}
\usepackage{amssymb}
\usepackage{pifont}
\usepackage{graphicx}
\usepackage{array}
\usepackage{caption}
\usepackage{ctable}
\usepackage[font=footnotesize]{subfig}
\newcommand{\rulesep}{\unskip\ \vrule\ }
\journal{Computer Vision and Image Understanding}

\begin{document}

\begin{frontmatter}



\title{Harnessing Noisy Web Images for Deep Representation}

\author{Phong D. Vo, Alexandru Ginsca, Herv\'e Le Borgne, Adrian Popescu}
\address{Vision \& Content Engineering Laboratory \\CEA LIST, France}

\author{}

\address{}

\begin{abstract}
The keep-growing content of Web images is probably the next important data source to scale up deep neural networks which recently surpass human in image classification tasks. The fact that deep networks are hungry for labeled data limits themselves from extracting valuable information of Web images which are abundant and cheap. There have been efforts to train neural networks such as autoencoders with respect to either unsupervised or semi-supervised settings. Nonetheless they are less performant than supervised methods partly because the loss function used in unsupervised methods, for instance Euclidean loss, failed to guide the network to learn discriminative features and ignore unnecessary details. We instead train convolutional networks in a supervised setting but use weakly labeled data which are large amounts of unannotated Web images downloaded from Flickr and Bing. Our experiments are conducted at several data scales, with different choices of network architecture, and alternating between different data preprocessing techniques. The effectiveness of our approach is shown by the good generalization of the learned representations with new six public datasets. 
\end{abstract}


\end{frontmatter}


\section{Introduction}\label{sec:intro} 
\noindent For a long time the vision community has been striving for the quest of creating human-like intelligent systems. Recently the resurgence of neural networks~\cite{DBLP:conf/ijcai/Hinton05,DBLP:journals/neco/HintonOT06,bengio2007greedy} has first led to a revolution in computer vision, for example~\cite{DBLP:journals/nn/CiresanMMS12,DBLP:conf/nips/KrizhevskySH12,DBLP:journals/corr/SimonyanZ14a,DBLP:journals/corr/SzegedyLJSRAEVR14,razavian2014cnn}, and then quickly provoked to other areas including reinforcement learning~\cite{DBLP:journals/corr/MnihKSGAWR13}, speech recognition~\cite{DBLP:conf/icassp/GravesMH13}, and natural language processing~\cite{DBLP:conf/nips/MikolovSCCD13}. For the most part those neural network models are supervised ones which require lots of labeled training data hence pose scalability challenges. This paper studies an alternative to train deep neural networks using massive amount of unannotated Web images.

Convnets have been well-known for the excellent generalization of its learned representation and being widely acknowledged as the \emph{de facto} representation learning method. A convnet is no more than an end-to-end feature mapping, i.e. starting from raw pixel intensities and then through many hidden layers of different types a robust representation can be learned. At the top of the network there is often a layer representing some loss function which is specific to each problem.  Giryes et al.~\cite{DBLP:journals/corr/GiryesSB15} proved that under random Gaussian weights deep neural networks are distance-preserving mappings with a special treatment for intra- and inter-class data.  

What makes convnets special is that it learns distributed representation\footnote{ This mean one concept is represented by multiple neurons and each neuron participates in the representation of more than one concept}. Distributed representation is indeed much more expressive than a local representation due to its compactness in term of lesser hidden units~\cite{DBLP:conf/nips/DelalleauB11} and much more regions of linearity~\cite{DBLP:conf/nips/MontufarPCB14}. Theoretical justifications of deep networks as a class of universal approximators~\cite{DBLP:journals/mcss/Cybenko92,DBLP:journals/nn/HornikSW89}. A more recent work \cite{DBLP:conf/icml/AnBB15} proved that a two-layer rectifier network can make any disjoint data linearly separable. 

While distributed representation is commonly present in many deep networks, convolutional and pooling layers, which are exclusive in convnet, are known to shift-invariance~\cite{DBLP:journals/corr/abs-1301-3537} and local context preservation. In fact convolutional layers are crucial for convnet to obtain better representation than other deep networks such as stacked auto-encoders, i.e. compare reported results in \cite{DBLP:conf/icml/LeRMDCCDN12} and \cite{DBLP:conf/nips/KrizhevskySH12}. Importantly convnets go beyond the i.i.d (independent identical distribution) assumption where their  inner representation is highly transferable to related tasks. For example the convnet model trained for image classification~\cite{DBLP:conf/nips/KrizhevskySH12} can be used as a feature detector~\cite{DBLP:journals/corr/RazavianASC14} for object detection~\cite{DBLP:conf/cvpr/ErhanSTA14}, image segmentation~\cite{DBLP:journals/corr/LongSD14}, and image retrieval~\cite{DBLP:conf/eccv/BabenkoSCL14,DBLP:conf/cvpr/WangSLRWPCW14}. 

Representation learning has been continually pursued by unsupervised methods such as auto-encoders~\cite{hinton2006reducing}, deep belief nets~\cite{DBLP:journals/neco/HintonOT06}, and sparse encoding~\cite{poultney2006efficient}, however from our viewpoint representation learning should combine advantages of both supervised and unsupervised regimes. Unsupervised learning alone is lack of a strong data prior. Since label information of training data is unavailable in an unsupervised setting, the objective function of an unsupervised network uses reconstruction loss, for example~\cite{hinton2006reducing}. This loss concerns too much on redundant image details, i.e. it tries to reconstruct at much at possible input images at pixel level hence  makes it less capable of generalizing discriminative features from visual variations. Supervised learning on the contrary can access to labels of training data thus is better guided. By minimizing the classification loss, supervised training helps pruning unnecessary details and magnifying discriminative features. 

Our perspective is also shared in~\cite{DBLP:conf/nips/DosovitskiySRB14,DBLP:journals/corr/Valpola14,DBLP:journals/corr/RasmusVHBR15}. Interestingly we find \cite{DBLP:conf/nips/DosovitskiySRB14}  was the pioneer to train general feature detectors using supervised convnets combined with artificially generated training data labeling information is instance-based rather than class-based. The learned representation therefore is quite robust and outperforms other unsupervised representation learning methods. Although the method proposed in \cite{DBLP:conf/nips/DosovitskiySRB14} is limited to small images of dimension $32\times 32$, it demonstrates the validity of training convnets for representation learning.

Inspired by \cite{DBLP:conf/nips/DosovitskiySRB14} we explore the approach of training large-scale convnets under supervised regime using weakly labeled data. Notice that this perspective is different from a known way~\cite{DBLP:journals/neco/HintonOT06} of combining unsupervised and supervised learning where the unsupervised part initializes weights which is then followed by supervised fine-tuning. Prior to deep learning, there have been some works~\cite{DBLP:journals/tmm/UlgesWB11,DBLP:journals/pami/WangHF12} that uses images harvested from Internet and photo sharing sites such as Flickr to train scalable image classifiers. However there is absent a thoroughly empirical analysis on the effectiveness of using noisy Web images to train deep networks. This is where our work comes into the context. 

Working with Web images comes with both pros and cons where images are cheap and abundant but very noisy. The advantage of Web images easily satisfies the data-hungry property of convnet; our only concern is the tolerance of convnet against noises. Lately we learned that convnet is surprisingly noise tolerable. In~\cite{sukhbaatar2014learning} and soon followed by \cite{xiao2015learning} studied several solutions to train deep convolutional networks as classifiers under noisy condition. In their works training data are assumed to contain mislabeled images so that probabilistic frameworks are proposed to estimate conditional mislabeling probabilities. Finally those probabilities are integrated into extra label noise layers placed at the top of convnet in order to improve posterior predictions. Different from \cite{sukhbaatar2014learning,xiao2015learning} we are rather interested in building a robust representation for general purposes from noisy data. Our experiments shown that even without any of special treatment of noisy images, convnet already performs very well. We aim to improve further this performance, not just limited in few specific cases but across a variety of domains. 

Our contribution is twofold. First, we train convnets using noisy and unannotated Web images retrieved from the image search engine Bing and the photo sharing network Flickr. Experiments are scaled from a small image collection of  hundred concepts and 400K images to a larger collection with a thousand concepts with 3.14 million images. In both scales the learned representations provide very generalized features that lead to promising accuracies on many classification datasets. Second, we convey image reranking techniques to remove noises from training data and train convnets of deeper architectures. Results show that the proposed techniques help improving classification results significantly. The best of our performance outperforms CaffeNet and close the gap with Vgg-16~\cite{DBLP:journals/corr/SimonyanZ14a}. 

In the remainder of this paper, we present data collection procedures in Section~\ref{sec:datacollection} and methods in Section~\ref{sec:method}. Section~\ref{sec:expt} presents experiment results and Section~\ref{sec:conclu} concludes the paper.

\section{Image Collections}\label{sec:datacollection}
\noindent Data acquisition takes an important role in our study. Data sources of Web images are so vast and diverse that it is better not to rely on a single source. However image crawling often comes up with a price and user privacy problems. Social network platforms like Facebook and Instagram either have strict privacy policies or simply do not provide image search API. Nevertheless for research Flickr always comes first with abundant data source and free of charge. Flickr photos are so diverse that they are not biased toward any of particular themes. Some Flickr photos are organized into groups and galleries which turns out that image search on Flickr gives quite relevant results. Indeed many public datasets have adopted Flickr as one of their principal sources, for example VOC Pascal challenges~\cite{Everingham10} and ImageNet~\cite{ILSVRC15}. 

Recently Flickr released the YFCC collection of 100 millions images and videos for research purpose. While this collection is very useful to experiment with unsupervised learning and data mining, we stitch to the most general approach where images of any concept or class can be retrieved using search engines. Retrieved images therefore truly reflect challenges caused by noisy images in practice. 

Because we want to compare our approach with the standard supervised approach that uses clean images from ImageNet, our collections are based on Wordnet\footnote{The lexical database of English organized like a thesaurus in which words having similar meanings (synonyms) are grouped into \emph{synsets}; a synset expresses a concept}. Given a synset in Wordnet, its synonyms are used as keywords to retrieve images. Downloaded images are not subjected to manual screening except duplicating images removal. Data imbalance between synsets is avoided by setting an equal number of images per synset. 

Besides Flickr we use Bing as a  complementary source. Using more data sources also prevent our collections from  biased. Thanks to the Bing Azure API we can freely download up to 250,000 images per account per month. Notice that we use a text query without specifying any of visual filter. Manual examination on downloaded images from Bing gives us a sense that Bing images are quite noisy, and at some extent they are much more noisy than Flickr images.

\begin{figure}[!t]
\centering
\subfloat[Images from Bing]{\includegraphics[width=0.3\linewidth]{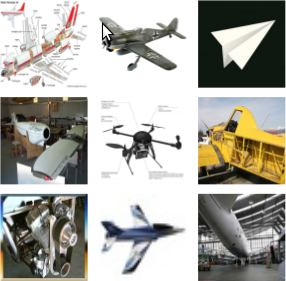} \rulesep \includegraphics[width=0.3\linewidth]{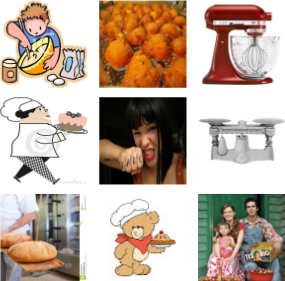} \rulesep \includegraphics[width=0.3\linewidth]{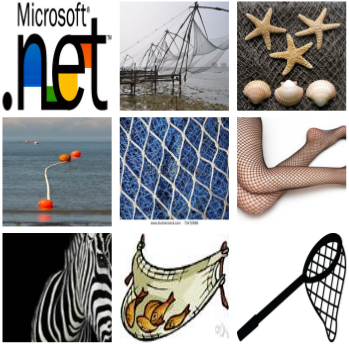}}\hspace{0.6cm}
\subfloat[Images from Flickr]{\includegraphics[width=0.3\linewidth]{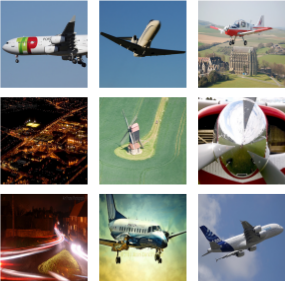} \rulesep \includegraphics[width=0.3\linewidth]{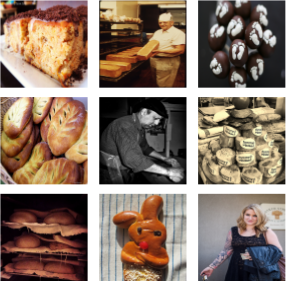} \rulesep \includegraphics[width=0.3\linewidth]{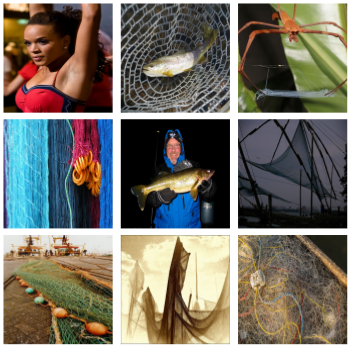}}\hspace{0.6cm}
\subfloat[Images from ImageNet]{\includegraphics[width=0.3\linewidth]{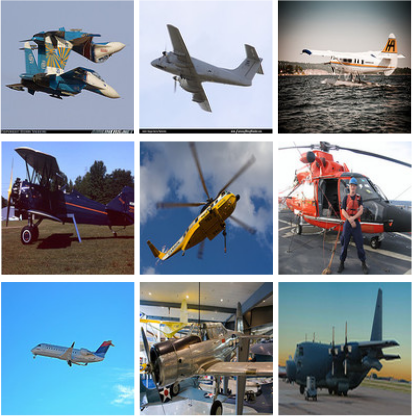} \rulesep \includegraphics[width=0.3\linewidth]{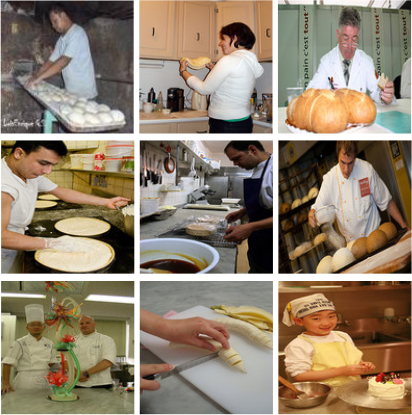} \rulesep \includegraphics[width=0.3\linewidth]{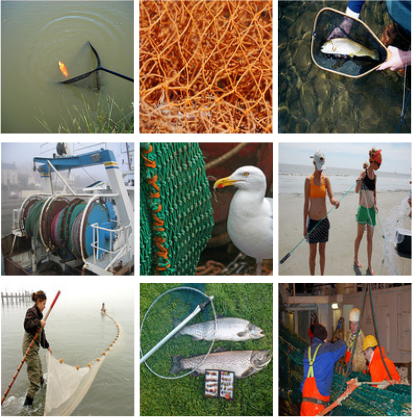}}
\caption{Some examples of three synsets \emph{airframe}, \emph{baker}, \emph{fishnet}, sampled from two Web data sources (a) Bing, (b) Flickr, and a human-annotated source ImageNet in (c). Data bias is obvious among the three sources. Noise level also varies from source to source; for instance Bing's \emph{fishnet} contains not only fishnet related images but also polyseme (legging shocks), texture related images (zebras), and even a logo of Microsoft .NET. Noisy examples from Flickr, on the other hand, are somewhat related to the synset query. For instance, there are aerial views in the synset \emph{airplane}, or there are cake images in the synset \emph{baker}. }
\label{fig:db}
\end{figure}
Depicted in Fig.~\ref{fig:db} are some examples from our Web image collections. At a first glance, these examples expose both high intra-variance per category and inter-variance between data sources. While the former is unavoidable and has to be reduced by means of image reranking techniques, the effect of the latter is unknown. At a closer look, Bing seems to have more documentary images and diagrams while Flickr has more personal photos with better aesthetic quality. This distinction is not difficult to explain. Bing Search is based on text so that images with rich accompanying texts are well indexed so that they appear in top results. On the contrary images of Flickr are uploaded, tagged, and organized by users; some of images are very relevant to  specific topics but those topics may irrelevant to search queries. Flickr mostly contains natural photos so that it is unlikely to contains cartoons or sketches.

Multiple data sources bring both advantages and challenges. In the one hand it  improves diversity. In the other hand it may reduce intra-class consistency. To give a final conclusion, we trained two convnets that either merely use Flickr images or mix Bing and Flickr images; classification results on some third-party datasets show that using more than one data source leads to better generalization. As a result experiments in the rest of this paper use both Flickr and Bing images. 

Studying the effect of noisy data onto representation learning should be done in different problem scales because results may drastically change as more noisy data take part in. We conduct experiments at two scales: the small collection of 100 synsets and the large collection of 1000 synsets. With the small collection we can quickly test to find out optimal hyperparameter settings; doing this on a large-scale model is very time-consuming and expensive. Once a good setting has been found, it will be is applied onto the large-scale problem. The two collections are described in the following.
\subsection{Flickr-Bing 100 (FB-s)} 
This collection consists of 100 synsets randomly sampled from WordNet. The number of images per synset ranges from 3000 up to 5000 images. Out of the total 416,000 images, Flickr and Bing contribute 67 \% and 33 \% respectively. Using the same set of 100 synsets, we create the baseline collection (IN-s) whose images are sampled from ImageNet; each synset contains approximately 1000 images thus IN-s has 100,000 images in total. The baseline dataset IN-s is used as the training data for fully supervised convnets. Evaluating relative performance of convnets trained from FB-s and IN-s will reveal how good the weak-supervised approach perform. 
\subsection{Flickr-Bing 1000 (FB-l)} 
This collection consists of 1000 synsets which are officially used in ILSVRC image classification challenges~\cite{ILSVRC15}. Each synset has approximately 3000 images thus the total number of images in FB-l is about 3.12 million images in which Flickr and Bing contribute 70\% and 30\% respectively. In fact there is no special reason to prevent us from using other synsets than those included in the ILSVRC challenges. Adopting the synset set of ILSVRC challenge henceforth does not reduce the generality of the approach and furthermore we can easily compare our results with existing works. As a result we do not need to prepare another baseline dataset as done with FB-s.

\section{Method}\label{sec:method}
\noindent Our method consists of two stages: i) partly remove noisy images and outliers from the collection, ii) train convnets with the refined collection. Image reranking is used in the first stage in order to rerank relevant images (clean data) to be at the top while pushing irrelevant images (a.k.a noises and outliers) out of the top list. For reranking is just a preprocessing step, it is regarded as being helpful if the learned representation produces better performance in a classification task. 

Given an arbitrary synset, let us call $\mathcal{L}=\{ (\mathbf{x}_i, y_i) \}_{i=1}^{m}$ the set of labeled examples and $\mathcal{U} = \{ \mathbf{z}_j \}_{j=1}^{n}$ the set of unlabeled examples which can be Web images in our context. Here, $\mathbf{x}_i$ and $\mathbf{z}_j$ are the vectorial representations of the corresponding labeled instance $i$ and unlabeled instance $j$; we also assume that $m \ll n$ to emphasize the necessity of semi-supervised reranking methods where examples are scarce. A reranking algorithm aims to select a subset $\mathcal{S} \subset \mathcal{U}$ such that $\mathbf{z}_k \in \mathcal{S}$ are more relevant to at least one among $\mathbf{x}_i \in \mathcal{L}$, than to $\mathbf{z}_j\in\mathcal{U} \backslash \mathcal{S}$.

\subsection{Cross-Validation (CV)} 
\noindent This technique splits $\mathcal{U}$ into $K$ equal disjoint subsets. Each subset is scored by a binary SVM classifier~\cite{cortes1995support} trained on the rest $(K-1)$ subsets as positive samples and other 10K images as negative samples. The latter can be obtained with ease: either subsampling images of synsets which are not relevant to the synset of interest. Iterating $K$ times gives us exactly one prediction for every data point in $\mathcal{U}$. Samples with negative scores are listed as noise thus rejected. The hyperparameter $K$ is manually chosen; increasing $K$ make lesser images classified as noise.

\subsection{Kernel Mean Matching (KMM)} 
This is a semi-supervised technique~\cite{DBLP:conf/nips/HuangSGBS06} that reweights unlabeled data  $\mathbf{z}_i \in \mathcal{U}$  w.r.t labeled data $\mathbf{x}_i \in \mathcal{L}$ such that the (weighted) arithmetic means of the two sets are approximately equal, i.e. $\sum \mathbf{x}_i/m \approx \sum \alpha_i \mathbf{z}_i/\left( \sum \alpha_j \right)$. If $\alpha_i \approx 0$ then $\mathbf{z}_i$ is considered as noise. The optimal $\boldsymbol{\alpha}^*$ is the solution of the following convex quadratic program
\begin{equation}
\text{arg }\underset{\begin{subarray}{c} \boldsymbol{\alpha} \succeq 0 \\ \boldsymbol{\alpha}' \mathbf{1} \approx n  \end{subarray}}{\text{min}} \frac{1}{2} \left\Vert \frac{1}{m} \sum_{i=1}^m \mathbf{x}_i - \frac{1}{n} \sum_{j=1}^n \alpha_j \mathbf{z}_j \right\Vert^2.
\label{eq:kmm}
\end{equation} 
Notice that Eq.~\ref{eq:kmm} operates directly on the input features without passing them through any nonlinear mapping. Using linear KMM therefore is really fast. Eq. (\ref{eq:kmm}) is convex and can be expressed as the canonical quadratic form as follows:
\begin{equation}
\begin{array}{cl}
\underset{\boldsymbol{\alpha}}{\text{arg min}} & \frac{1}{2} \boldsymbol{\alpha}' \left( \mathbf{Z}' \mathbf{Z} \right) \boldsymbol{\alpha} - \frac{n}{m} (\mathbf{Z}' \mathbf{X} \mathbf{1})' \boldsymbol{\alpha} \\
\text{s.t} & 0 \leq \boldsymbol{\alpha} \leq B \\
 & m(1-\epsilon) \leq \boldsymbol{\alpha}' \mathbf{1} \leq m(1 + \epsilon)
 \label{eq:kmm2}
\end{array},
\end{equation} where the first constraint defines a scope bounding discrepancy between the distributions of $\mathcal{L}$ and $\mathcal{U}$. The bigger $B$ is, the lesser number of points $\mathbf{z}_i$'s are highly re-weighted. As value of $B$ approaches zero, an unweighted solution is obtained. The second constraint ensures the measurement $\boldsymbol{\alpha}(x)P(x)$ is close to a probability distribution (for further details see~\cite{DBLP:conf/cvpr/ChuTC13}).
\subsection{Transductive Support Vector Machine (TSVM)}
\noindent Proposed by \cite{DBLP:conf/sigir/SindhwaniK06}, TSVM uses both labeled data  $\mathcal{L}$ and unlabeled data $\mathcal{U}$ to infer a decision function. In our context it is the noise removal function $\mathbf{w}$. According to the setting of TSVM, $|\mathcal{L}|$ could be much smaller than $|\mathcal{U}|$, which perfectly fits to our context. To find  $\mathbf{w}$, the following quadratic program must be iteratively solved
\begin{equation}
\text{arg }\underset{\mathbf{w},\{t_j\}}{\text{min}} \frac{1}{2} \left\Vert \mathbf{w} \right\Vert_2^2 + \frac{\alpha}{m} \sum_{i=1}^m \ell(y_i \mathbf{w}' \mathbf{x}_i) + \frac{\beta}{n} \sum_{j=1}^n \ell(t_j \mathbf{w}' \mathbf{z}_j) \text{ s.t } \frac{|\{t_j^+\}|}{|\{t_j^-\}|}=\rho,
\label{eq:tsvm}
\end{equation} where hyperparameters $\alpha$ and $\beta$ control the influence of labeled data $\left \{{\mathbf{x}_i}\right \}$ and unlabeled data $\left \{{\mathbf{z}_j}\right \}$ on the classifier $\mathbf{w}$; the loss $\ell(\cdot)$ penalizes the predicted labels $\hat{t}_j = \text{sign}(\mathbf{w}' \mathbf{z}_j)$ and $\hat{x}_i = \text{sign}(\mathbf{w}' \mathbf{x}_i)$ w.r.t its temporary label $t_i$ and groundtruth $y_i$ respectively. Because $\mathbf{x}$ and $\{ t_i \}$ are coupled by the second loss term, Eq.~\ref{eq:tsvm} therefore is non-convex and could be solved by alternating minimization. In particular, $\{t_j\}$ is the set of temporary labels of $\{ \mathbf{z}_j \}$ during optimization, i.e. $\{t_j\}^{(\tau)}$ is assigned by $\mathbf{w}^{(\tau)}$ at iteration $\tau$-th.  The optimization process terminates if either $\{ t_j \}^{(\tau)} \equiv \{ t_j \}^{(\tau+1)}$ or the maximum number of iteration is reached. As a result an unlabeled point $\mathbf{z}_j$ is classified as noise if $t_j=-1$. To avoid a  trivial solution where all of $\{\mathbf{z}_j\}$ falls into either positive or negative side, the ratio of positive labels $\{ t_j \}_+$ and negative labels $\{ t_j \}_-$ is set to $\rho$, i.e. $0<\rho<1$. 

\subsection{Convnet Architectures}\label{sec:deeper}
\noindent With millions of learnable parameters, dozens of hyperparameters and network topology, finding a neural net architecture appropriate for a task is more an art than science. Fortunately convnet architecture is somewhat constrained by feedforward learning and relative order of layer types. Currently a handful of convnets performs seamlessly, for example AlexNet~\cite{DBLP:conf/nips/KrizhevskySH12} or its slight variation CaffeNet~\cite{jia2014caffe}, vgg  nets~\cite{DBLP:journals/corr/SimonyanZ14a}, PreLU nets~\cite{he2015delving}, Google net~\cite{DBLP:journals/corr/SzegedyLJSRAEVR14}. In our experiments we choose Alex net~\cite{DBLP:conf/nips/KrizhevskySH12} as a starting point and then try to increase network's depth with a modified structure. 

In most cases increasing the depth of convnets depth leads to performance gain, for example~\cite{DBLP:journals/corr/SimonyanZ14a,he2015delving,DBLP:journals/corr/SzegedyLJSRAEVR14}. However, \cite{DBLP:journals/corr/SimonyanZ14a} experiences impossibility of training a convnet with more than 16 layers which can be overcame by layer pre-initialization~\cite{DBLP:journals/corr/SimonyanZ14a, romero2014fitnets}. We encounter this problem at a lesser number of layers, for instance we cannot train a 16-layer convnet using FB-l as training data. Since the major difference between our experiments and other works, for example \cite{DBLP:journals/corr/SimonyanZ14a}, is the use of weakly labeled and noisy data. Rather of initializing intermediate layers with pre-trained weights, some technical modifications can resolve this problem.

A foremost factor is the minibatch size of stochastic gradient descent (SGD) algorithm used to train the network. In practice convnets can be trained with a very small batch size up to 16 images per batch (with lowered learning curve), this is no longer true in our cases in which training data are heavily corrupted by noises. We found that with a too small batch size a convnet like Vgg-16 could not decrease training loss even after thousands of iterations. Our conjecture is that the  fluctuation of of gradient directions (due to noises) between subsequent batch subsamplings slowdowns learning speed of convnets.

Another factor is the size of convolutional kernels. Recent findings~\cite{DBLP:journals/corr/SimonyanZ14a} suggests that deeper convnets with small kernel sizes such as $3\times 3$ tends to improve generalization. Our results on the contrary shows that medium kernel sizes tends to work better for Web images. While we have not figured out evidences of the association between image appearances and kernel sizes, a plausible explanation can be based on over-flourishing appearances of Web images. Such images come from a variety of contexts and objects contained in them may occur at any scales and styles such as cartoons, diagrams, sketches.

Based on the observations above, we propose the 13-layer network architecture called FBNet as shown in Table~\ref{table:protos}. Our network is very much alike Vgg-16 net  except that it just has 4 max-pooling layers and the kernel dimension of the first convolutional layer is double size of Vgg-16 net ($3 \times 3$) and half of CaffeNet ($11 \times 11$). While we are interested in increasing further the depth of our convnet, limited time and computational resources give us maximum 13 layers, which fits to a single GTX Titan-Z 12GB memory. Each 100 iterations takes around 8 minutes for the batch size 196; it takes about 3 weeks for a full training from scratch.

An alternative way to attain more depth is to try  GoogleNet~\cite{DBLP:journals/corr/SzegedyLJSRAEVR14}. This architecture is deeper than Vgg-16 but still consumes slightly less memory and even run faster than Vgg-16. We use this convnet without modification and train from scratch on Web data; on average it takes 2.5 minutes for each 100 iterations run with the mini-batch size 128. A full training stops after 1.2 million iterations.
\begin{table}
\small
\begin{centering}
\begin{tabular}{l|lll}
 Layer & CaffeNet & vgg16 & FBNet \tabularnewline
\hline
conv & $11\times 11, 96, /4$ & $3\times 3, 64$ & $7\times 7, 96, /2$ \tabularnewline
 & & $3\times 3, 64$ & \tabularnewline
 maxpool & $3\times 3, /2$ & $2\times 2, /2$ &  \tabularnewline
 \hline
conv & $5\times 5, 256$ & $3\times 3, 128$ & \tabularnewline
 & & $3\times 3, 128$ & \tabularnewline
 maxpool & $3\times 3, /2$ & $2\times 2, /2$ & $2\times 2, /2$ \tabularnewline
 \hline
conv & $3\times 3, 384$ & $3\times 3, 256$ & $3\times 3, 256$ \tabularnewline
 & & $3\times 3, 256$ & $3\times 3, 256$ \tabularnewline
 & & $3\times 3, 256$ & $3\times 3, 256$ \tabularnewline
 maxpool & & $2\times 2, /2$ & $2\times 2, /2$ \tabularnewline
 \hline
conv & $3\times 3, 384$ & $3\times 3, 512$ & $3\times 3, 512$ \tabularnewline
 & & $3\times 3, 512$ & $3\times 3, 512$ \tabularnewline
 & & $3\times 3, 512$ & $3\times 3, 512$ \tabularnewline
 maxpool & & $2\times 2, /2$ & $2\times 2, /2$ \tabularnewline
 \hline
conv & $3\times 3, 256$ & $3\times 3, 512$ & $3\times 3, 512$ \tabularnewline
 & & $3\times 3, 512$ & $3\times 3, 512$ \tabularnewline
 & & $3\times 3, 512$ & $3\times 3, 512$ \tabularnewline
 maxpool & $3\times 3, /2$ & $2\times 2, /2$ & $2\times 2, /2$ \tabularnewline
 \hline
 fc6 & \multicolumn{3}{c}{4096} \tabularnewline
 fc7 & \multicolumn{3}{c}{4096} \tabularnewline
 fc8 & \multicolumn{3}{c}{1000} \tabularnewline
\end{tabular}
\par\end{centering}
\caption{The feedforward architectures of two reference convnets versus ours (rightmost column).}
\label{table:protos}
\end{table}
\section{Experiment Setup}\label{sec:expt}
\begin{figure}[!t]
\centering`'
\subfloat[SUN 397]{\label{fig:sun397}\includegraphics[width=0.75\linewidth]{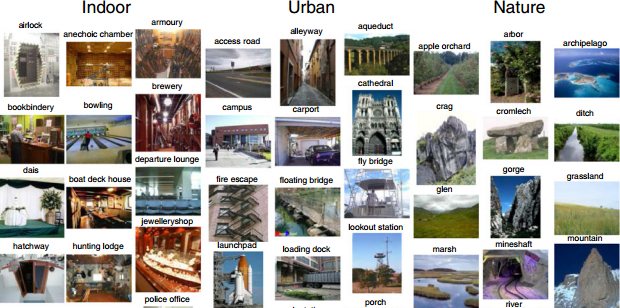}} \\
\subfloat[MIT Indoor 67]{\label{fig:indoor67}\includegraphics[height=0.4\linewidth]{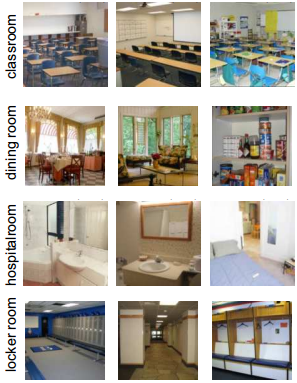}}\,
\subfloat[Caltech 256]{\label{fig:caltech256}\includegraphics[height=0.4\linewidth]{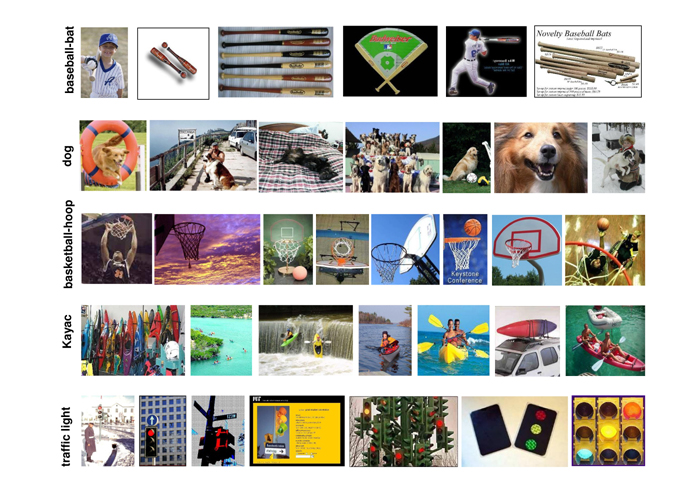}} \\
\subfloat[Pascal VOC 2007]{\label{fig:indoor67}\includegraphics[height=0.27\linewidth]{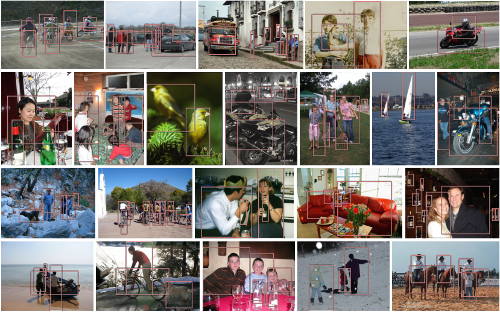}}\,
\subfloat[Action 40]{\label{fig:action40}\includegraphics[height=0.25\linewidth]{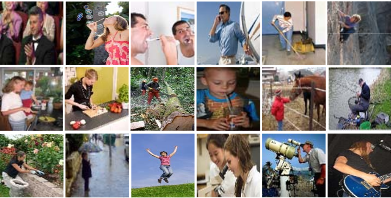}}
\label{fig:datasets}
\caption{MIT Indoor 67 and SUN 397 are about static scenes; Caltech256 contains object-centric images; Pascal VOC 2007 and Action 40 focus on objects and people in context.}
\end{figure}
The representation learned from Web images are evaluated on public datasets with various themes: indoor scenes MIT67~\cite{DBLP:conf/cvpr/QuattoniT09}, a variety of outdoor and street scenes SUN397~\cite{DBLP:conf/cvpr/XiaoHEOT10}, human actions Action40~\cite{DBLP:conf/iccv/YaoJKLGF11}, object categories Caltech256~\cite{griffin2007caltech}, objects in context VOC07~\cite{Everingham10}, three fine-grained datasets of flower species  Oxford102~\cite{Nilsback08}, dog species StandfordDogs~\cite{}, bird species CUB-200~\cite{WelinderEtal2010}, and one fine-grained dataset of car brands and models StanfordCars~\cite{krause20133d}. Mean accuracy is used to evaluate all of datasets except of VOC07 which is using mean average precision (mAP). 
\begin{table}
\small
\begin{centering}
\begin{tabular}{l|llll}
 Reranking & n/a & cvsvm & kmm & tsvm \tabularnewline
 \hline 
 FB-s ($\times 10^3$) & 406.5 & 291 & 81.3 & 98.9 \tabularnewline
 FB-l ($\times 10^6$) & 3.14 & 2.52 & 1.44 & 2.03 \tabularnewline
\end{tabular}
\par\end{centering}
\caption{The sizes of the small collection FB-s (100 synsets) and the large collection FB-l (1000 synsets) before (1st column) and after reranking (2nd, 3rd, 4th columns). Notice that semi-supervised reranking algorithms are more strict in filtering noisy images than the unsupervised \emph{cvsvm}. n/a means reranking not applied. }
\label{table:numimgs}
\end{table}
\subsection{Image Reranking}\label{sec:reranksetting}
\noindent Reranking algorithms require input features of images which are computed by some feature extraction method. In the case of FB-s, the CaffeNet distributed along with the Caffe toolkit~\cite{jia2014caffe}, which is pre-trained on ILSVRC'12 training data of 1000 categories, is used as a feature extractor. In the case of FB-l, we use the CaffeNet trained on FB-l itself as a feature extractor. In other word we apply the so called \emph{self-reranking} technique:  i) to train a convnet from Web images without reranking preprocessing, ii) use that convnet as a feature extractor back to the images used to train it, iii) apply reranking methods to remove noise and outliers. 

The feature extractor runs as follows. Images are forwarded from input layer and  4096-dimensional feature vectors can be extracted at the fully connected layer \texttt{fc7}; these features are normalized with $L_2$ scheme before fed into reranking algorithms. Notice that among the three reranking algorithms, \emph{cvsvm} is the only one that do not require labeled examples. Therefore we have to manually annotate a tiny set of examples for two semi-supervised methods \emph{kmm} and \emph{tsvm}. In particular $m=10$ labeled examples per synset are annotated, which means 1,000 and 10,000 examples of FB-s and FB-l are given to reranking algorithms. Other hyperparameter settings include:
\begin{itemize}
\item \emph{cvsvm}: liblinear is used to run its SVM sub-problems with linear kernel and $C=1$; the number of folds $K=5$; $K>5$ tends to accept more noisy images as clean.
\item \emph{kmm}: the quadratic equation is solved using CVXOPT; $B=5$ tends to balance between the need of more training images (decreasing $B$) versus noise removal (increasing $B$).
\item \emph{tsvm}: we reuse the svmlin code released by its author~\cite{DBLP:conf/sigir/SindhwaniK06}; $\rho$ is set to $1000/n$ so that for each of synset, about 1000 images is selected as clean; $\alpha=1$ and $\beta=1e-4$ emphasize the importance of labeled over unlabeled images. 
\end{itemize}
\subsection{Visualizing Reranking Results}\label{sec:vizrerank}
\begin{figure*}[!th]
\centering
\subfloat[\emph{cvsvm}: Cross-Validated SVM]{\label{fig:cv} \includegraphics[width=.45\linewidth]{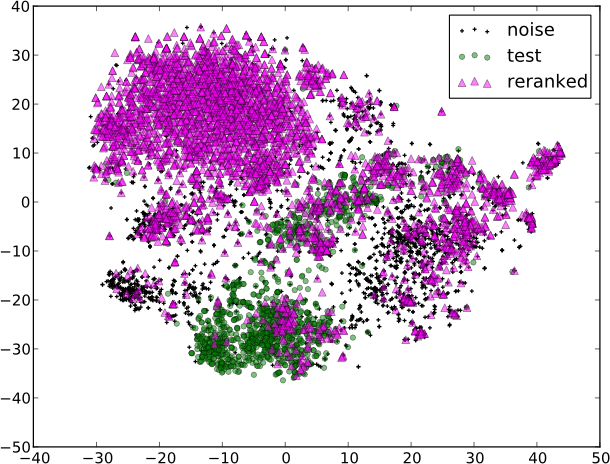}}\hspace{0.4cm}
\subfloat[\emph{tsvm}: Transductive SVM]{\label{fig:tsvm} \includegraphics[width=.45\linewidth]{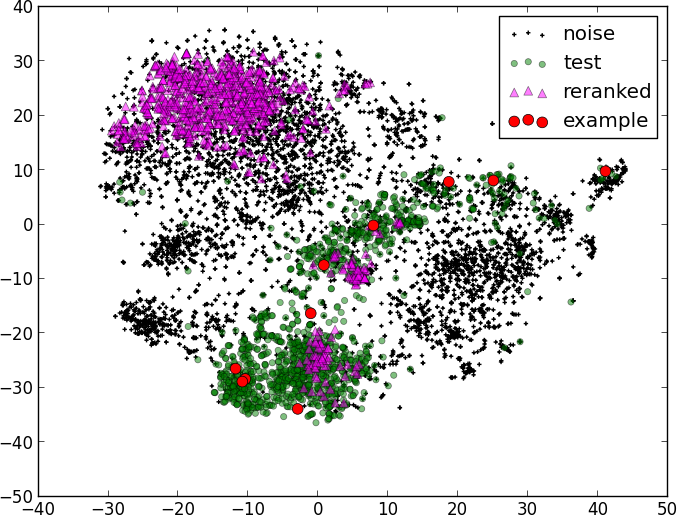}} \\
\subfloat[\emph{kmm}: Kernel Mean Matching]{\label{fig:kmm} \includegraphics[width=.9\linewidth]{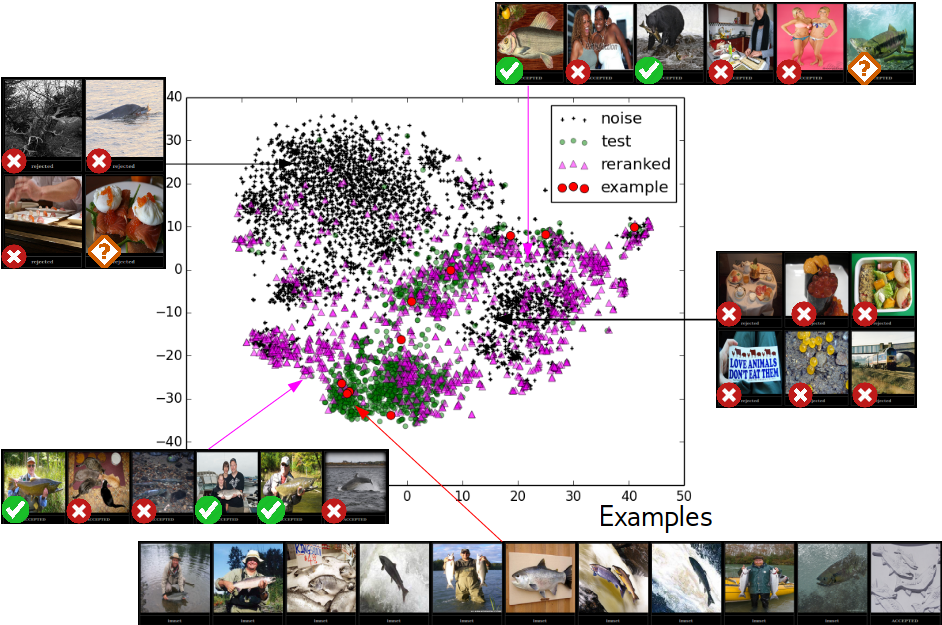}}
\caption{The behaviors of three reranking techniques(a)  \emph{cvsvm}, (b) \emph{kmm}, and (c) \emph{tsvm} on the synset \emph{salmon}. Red and green dots denote ImageNet images while black and pink dots denote FlickrBing images. Shown in green dots are clean images drawn from ImageNet, red dots denote examples given to the reranking algorithms, black dots denote noises determined by algorithms, and pink dots denote clean images determined by algorithms.}
\label{fig:tsnevis}
\end{figure*}
\noindent The effect of image reranking is not clear until we observe final classification results of the classifiers trained by reranked data. We can however observe  visualizations of the reranked data in order to have a sense of how algorithms handle data. We extract \texttt{fc7} features of images in the synset ``salmon'' of two small collections FB-s and IN-s, next run the dimensionality reduction method t-SNE~\cite{van2008visualizing}, then display the resulted 2D embedding in Fig.\ref{fig:tsnevis}. Notice that the reranking algorithms just operate on Web images (pink and black dots) and labeled examples (red); ImageNet images (green) are shown just for clarity purpose. 

We observe that the two distributions of Web images and ImageNet are not quite overlapped. This explains why \emph{cvsvm}, an unsupervised reranking method that does not use any examples, selects a large portion of Web images belonging to big clusters as clean data. On the contrary, semi-supervised methods \emph{tsvm} and \emph{kmm} favor in choosing Web data points surrounding examples. Here the difference between \emph{tsvm} and \emph{kmm} is clear: \emph{tsvm} takes into account both internal structure of Web images and provided examples, while \emph{kmm} disregards that structure of unlabeled images and keeps trying to match the empirical means of examples and the subset of clean images. At the end such  differences lead to different results of convnet training as we discuss in next sections.

\subsection{Convnet Training}\label{sec:convsetting}
\noindent Training convnets is time consuming, hence we try various settings of reranking first on the small collection FB-s to figure out working recipes and then apply onto the large collection FB-l. To know the effect of reranking, we train several convnets with respect to different data configurations. First, a convnet is trained from scratch using Web images without using any reranking method. Second, three convnets are trained from reranked images produced by the three methods respectively. As shown in Table~\ref{table:numimgs}, there are significant drops in term of number of images when applying reranking algorithms to the original Web collections. Because few training data tends to produce overfitting, the third configuration is to fine-tune three convnets (with respect to three reranking algorithms) based on the pre-trained model of the first configuration. 

For the FB-s collection we just use CaffeNet architecture while both CaffeNet and FBNet are used for the FB-l collection. Input image dimension is fixed to $227\times 227$, the optimization algorithm is SGD with momentum $0.99$; the learning rate $\eta=10^{-2}$ and drops by a magnitude of $1/\gamma=10$ after a step size of $10^4$ iterations; the maximum number of iterations is 450,000. The learning rate for fine-tuning is ten times less, i.e. $\eta=10^{-3}$, the corresponding step size is also shorter, which is from $3\times 10^3$ to $4\times 10^3$. A fine-tuning process is stopped after maximum 150,000 iterations.

\section{Results}\label{sec:fb100}
\noindent In this section we evaluate the generalization of the learned representations w.r.t Web image collections to unseen data. We present classification results considering convnets as end-to-end classifiers, however this serves as a reference and not the main purpose of our study. To test feature transferability of each trained convnet, we extract image features at layer \texttt{fc7} and train linear SVM classifiers for evaluation. 
\subsection{Results of FB-s}
\subsubsection{As End-to-End Classifiers}\label{sec:convcls}
\noindent We train caffe net for each of variant of the training data produced by individual reranking method. The accuracy is computed by comparing image groundtruth versus its softmax output at the last layer \texttt{fc8}. Results of classification accuracy on the test set of IN-s are summarized in Table~\ref{table:100}.  

Table~\ref{table:acc_INs} shows that training with reranked data turns out to be effective. This is true for the algorithm \emph{cvsvm} but neither \emph{kmm} nor \emph{tsvm}. We are curious about this and finally explain this as the consequence of overfitting. Semi-supervised reranking methods \emph{kmm} and \emph{tsvm} reject quite many images as noises so that the amount of training images is insufficient to train such a complex system like convnet. See Table~\ref{table:numimgs} to compare the number of images before and after reranking for each of method.
 
To achieve training convnets using less noisy data and avoiding overfitting, we apply \emph{two-step training}. In the first step We train from scratch a CaffeNet model using the entire Flickr and Bing images without reranking. Once finished its weights are used to initialize another CaffeNet model which will be trained (a.k.a fine-tuning) on  reranked data done by either \emph{cvsvm}, \emph{kmm}, or \emph{tsvm}. The second step spends considerably less training time, i.e. the maximum number of iterations may vary from 50,000 to 100,000. We also test another  \emph{two-step training} where the first step starts with FB-s/cvsvm data, and the second step uses either reranked data of \emph{kmm} or \emph{tsvm}.
 
Show in Table~\ref{table:acc_finetune_IN_s} are the classification results obtained by the two-step training and we observe significant improvement. As expected the results of convnets trained from reranked data FB-s/tsvm and FB-s/kmm are better than those of no reranking and unsupervised reranking \emph{cvsvm}. Noticeably the overall result is slightly better if the first step is trained by the full FB-s collection (compare the first 3 columns versus the last 2 columns). Among reranking methods, \emph{kmm} outperforms the rest probably due to its objective function of matching the empirical means of examples (drawn from the distribution that generates the test set IN-s) and the Web images.
 
When comparing the best result of the convnet trained from FB-s to the one trained from IN-s (see Table~\ref{table:acc_IN-s_vs_FB-s}), the latter obviously outperforms. This happens without surprise because the latter was trained and tested in the same data domain.
\begin{table}[!t]
\centering
\subfloat[In training from scratch, image reranking does not always improve accuracy. Training on the reranked set FB-s/cvsvm just  improve 0.7 point compared to using all images for training. Even worse training on the sets FB-s/kmm and FB-s/tsvm drop the performance. This may due to the numbers of images in FB-s/kmm and FB-s/tsvm reduce too much and lead to overfitting.]{
\begin{centering}
\begin{tabular}{l|cccc}
reranking & n/a & FB-s/cvsvm & FB-s/kmm & FB-s/tsvm \tabularnewline
\hline
mean accuracy & 53.6 & 54.3 & 48.9 & 51 \tabularnewline
\end{tabular}
\par\end{centering}
\label{table:acc_INs}}\\
\subfloat[Two-step training in which the second step uses the weights of the convnet trained in the first step as weight initialization. The training data in the second step is also ``cleaner.'' On the left three networks are initialized by the convnet trained on FB-s, then  use FB-s/cvsvm, FB-s/kmm and FB-s/tsvm respectively as training data for the second step. On the right, betworks are initialized by the convents trained on FB-s/cvsvm, then use FB-s/kmm and FB-s/tsvm respectively as training for the second step.]{
\begin{centering}
\begin{tabular}{l|ccc|cc}
initialized by & \multicolumn{3}{c}{CaffeNet on FB-s} & \multicolumn{2}{c}{CaffeNet on FB-s/cvsvm} \tabularnewline
 \hline
reranking & cvsvm & kmm & tsvm & kmm & tsvm \tabularnewline
 \hline\hline
mean accuracy & 55.3 & 58.7 & 57.7 & 58.4 & 57.6 \tabularnewline
\end{tabular}
\par\end{centering}
\label{table:acc_finetune_IN_s}}\\
\subfloat[Comparison between our best net versus the baseline, i.e. CaffeNet trained on the labeled data IN-s. Notice that the baseline is trained and tested on data drawn from ImageNet while our net is trained on Web images and tested on ImageNet.]{
\begin{centering}
\begin{tabular}{l|cc}
initialization & - & CaffeNet on FB-s \tabularnewline
 \hline
Training data & IN-s & FB-s/kmm \tabularnewline
\hline\hline
mean accuracy & 68.8 & 58.7 \tabularnewline
\end{tabular}
\par\end{centering}
\label{table:acc_IN-s_vs_FB-s}}
\caption{Mean classification accuracies of CaffeNet trained on FB-s with different choices of image reranking and weight initialization. In (a): Training from scratch, (b): \emph{two-step} training, (c): compare with the baseline. The test set is sampled from 100 ImageNet synsets and consists of 10,000 clean images.}
\label{table:100}
\end{table}
\subsubsection{Transferability}\label{sec:transfer}
\noindent We test on six public datasets. Images of each dataset are forwarded through the net and at the layer \texttt{fc7} we obtain 4096-dimensional features; applying $L_2$ normalization on these features and then we feed them to the training stage of one-vs-rest linear SVM classifiers. The choice of hyperparameters is made optimal per dataset per net. Evaluation protocol of each dataset is strictly followed. 

The results are shown in Table~\ref{table:transfer_100}. In Table~\ref{table:acc_IN-s_3rd} and Table~\ref{table:acc_IN-s_finetune_3rd} the relative performance between different nets are similar to Table~\ref{table:acc_INs} and Table~\ref{table:acc_finetune_IN_s}. In other word, the \emph{two-step} training scheme consistently improve the generalization of learned representations for within-domain and new domains. When comparing our best net against the CaffeNet trained on IN-s, it is surprising that our net outperforms with a large margin in 6 out of 6 datasets. We consider this as a promising signal for our approach but inferior performance of IN-s may due to its relatively small training data compared to FB-s.
\begin{table}[!t]
\centering
\subfloat[Evaluation on transferability of convnets trained from scratch.]{
\begin{centering}
\begin{tabular}{l|cccc}
reranking & n/a & cvsvm & kmm & tsvm \tabularnewline
\hline
voc07 & 57.9 & 57.9 & 48.5 & 54.3 \tabularnewline
oxford102 & 74.2 & 74.4 & 64.3 & 72.4 \tabularnewline
mit67 & 34.1 & 34 & 27.2 & 31.4 \tabularnewline
action40 & 37.9 & 38.2 & 29.1 & 35.6 \tabularnewline
caltech256 & 44.7 & 44.6 & 36.4 & 42 \tabularnewline
sun397 & 28.6 & 28.6 & 20.8 & 26.2 \tabularnewline
\end{tabular}
\par\end{centering}
\label{table:acc_IN-s_3rd}}\\
\subfloat[Evaluation on transferability of convnets trained in two-step. ]{
\begin{centering}
\begin{tabular}{l|lll|ll}
initialized by & \multicolumn{3}{c}{CaffeNet on FB-s} & \multicolumn{2}{c}{CaffeNet on FB-s/cvsvm} \tabularnewline
 \hline
reranking & cvsvm & kmm & tsvm & kmm & tsvm \tabularnewline
\hline\hline
voc07 & 58.2 & 57.5 &  58.8& 58.2 & 58.6 \tabularnewline
oxford102 & 74.3 & 75.4 & 77.7 & 75.3 & 76.1 \tabularnewline
mit67 & 34.6 & 33 & 38.3 & 34.9 & 36.3 \tabularnewline
action40 & 39.2 & 38.9 & 41.4 & 38.3 & 40 \tabularnewline
caltech256 & 44.6 & 44.5 & 46.7 & 44.7 & 46.7 \tabularnewline
sun397 & 28.7 & 29.1 & 31.1 & 28.7 & 30.7 \tabularnewline
\end{tabular}
\par\end{centering}
\label{table:acc_IN-s_finetune_3rd}}\\
\subfloat[Our best net versus the baseline.]{
\begin{centering}
\begin{tabular}{l|cc}
initialized by & - & CaffeNet on FB-s \tabularnewline
 \hline
Training data & IN-s & FB-s/tsvm \tabularnewline
\hline\hline
voc07 & 54.7 & 58.8 \tabularnewline
oxford102 & 72.4 & 77.7 \tabularnewline
mit67 & 31.9 & 38.3 \tabularnewline 
action40 & 36.8 & 41.4 \tabularnewline
caltech256 & 42.5 & 46.7 \tabularnewline
sun397 & 26.9 & 31.1 \tabularnewline
\end{tabular}
\par\end{centering}
\label{table:acc_IN-s_compare_3rd}}
\caption{Evaluation on transferability. In (a): convnets trained from scratch, (b): convnets trained in two steps, (c): compare with baseline.}
\label{table:transfer_100}
\end{table}
\subsection{Results of FB-l}\label{sec:fb1k}
\noindent Extending the experiment above at larger data scales is necessary to verify the scalability of our approach. In this section we repeat the experiments with the large collection FB-l of 1000 synsets. Based on previous results we select well performing nets for this new experiment so that we avoid wasting lots of training time on suboptimal configurations. 

Notice that we use \emph{self-reranking} to produce reranked images. In other word  self-reranking firstly trains a CaffeNet from raw Web images and then use this net as a feature extractor to compute features required by reranking algorithms. 
\subsubsection{As End-to-End Classifiers}
\noindent Shown in Table~\ref{table:classify1k} are classification accuracies of our convnets on the ILSVRC 2012 validation set. Compare with the previous results in Table~\ref{table:100}, this time our convnets perform poorly in which the best top-5 accuracy obtained by the convnet trained on FB-l/cvsvm is 23.8\%, a huge drop compared to 80.4\% of the convnet trained on IN-l. This means that when more data involved, the number of noisy images grow up and this leads to degraded classification results. The message of the results is also clear: to use a convnet as a good end-to-end classifier, having lots of clean labeled training data is a must. 
\begin{table}[!t]
\centering
\subfloat[The convnet trained on IN-l obtains absolutely superior performance. ]{
\label{table:classify1k}
\begin{centering}
\begin{tabular}{l|ccccc}
initialization & - & - & \multicolumn{3}{c}{CaffeNet on FB-s} \tabularnewline
\hline
training data & IN-l & FB-l & FB-l/cvsvm & FB-l/kmm & FB-l/tsvm \tabularnewline
\hline\hline
mean accuracy & 80.4 & 23.4 & 23.8 & 23.1 & 22.8 \tabularnewline
\end{tabular}
\par\end{centering}}\\
\subfloat[Transferring learned representation of ImageNet and Web images onto new domains. The results of the convnet trained on IN-l are better in almost all of the cases but convnets trained from Web images also follow closely. ]{
\label{table:transfer1k}
\begin{centering}
\begin{tabular}{l|ccccc}
initialization & - & - & \multicolumn{3}{c}{CaffeNet on FB-s} \tabularnewline
\hline
training data & IN-l & FB-l & FB-l/cvsvm & FB-l/kmm & FB-l/tsvm \tabularnewline
\hline\hline
voc07 & 75.3 & 74.4 & 74.7 & 71.0 & 72.5 \tabularnewline
oxford102 & 87 & 87.6 & 88.4 & 88.6 & 89.4 \tabularnewline
mit67 & 60.8 & 58.4 & 59.1 & 54.0 & 56.8 \tabularnewline
action40 & 60.2 & 55.9 & 56.5 & 56.1 & 56.3 \tabularnewline
caltech256 & 70.3 & 68.6 & 69.5 & 68.9 & 69.7 \tabularnewline
sun397 & 46.1 & 45.8 & 46.1 & 45.4 & 45.4 \tabularnewline
\end{tabular}
\par\end{centering}}
\caption{Classification accuracy of convnets trained on IN-l and FB-l w.r.t the test set of IN-l (a) and evaluation on transferring learned representation from ImageNet and Web images onto new domains (b).}
\label{table:fb-l}
\end{table}
\subsubsection{Transferability}\label{sec:transfer1k}
\noindent The rest question, as a reminder, which is also our question of interest: good generalization of a transferred representation requires abundant clean training data too? In the previous experiment with the small collection FB-s, the answer is no. For the large collection FB-l, the results are shown in Table~\ref{table:transfer1k}. Again the answer is no. The best of our results trained from 3.1 millions Web images are comparable to the reference model trained from 1.2 million ImageNet images. In three datasets VOC07, Caltech256, and SUN 297 our best performance obtained by the convnet trained from FB-l/cvsvm are mostly comparable to the reference model; it just performs worse than the reference at two datasets MIT67 and Action40.  

FB-l/cvsvm however slightly outperforms IN-l at Oxford102. This is interesting because unlike other datasets, the number of training images of Oxford102 is 1000 while the number of test images is 5000. This means each flower category just has 10 training images which is insufficient for convnet fine-tuning. The fact that the convnet trained on Web images generalizes better than the one trained on ImageNet on the Oxford102 dataset may imply a potential advantage of the proposed approach in domains with scarce data. 

In overall these results show that it is possible of obtaining competitive results by just using abundant amounts of unlabeled and noisy Web images. The results also show that reranking methods, especially the semi-supervised ones, have little effects and in some cases decrease the performance due to reducing the amount of training data via noise removal. In order to obtain again better results with reranking methods, more number of training images should be collected so that they are still abundant after rejecting noisy ones.
\subsection{Deeper Architectures}
\noindent According to recent empirical evidences the power of deep models depends on its depth. The 16-layer convnet Vgg-16~\cite{DBLP:journals/corr/SimonyanZ14a} and 22-layer convnet~\cite{DBLP:journals/corr/SzegedyLJSRAEVR14} have approached and surpassed human-level performance of image classification on ImageNet data. In our study we are also curious if a deeper convnet also learn a better representation with  noisy data. We train a 13-layer convnet, denoted as FBNet, with FB-l. Results are compared against CaffeNet and GoogleNet trained on the same dataset.

Training a very deep network is notoriously difficult. Back-propagating gradients from the loss layer make gradients vanished before reaching all layers below. Rectifier activation~\cite{DBLP:journals/jmlr/GlorotBB11} seems to solve this problem; furthermore training a very deep network requires a careful weight initialization~\cite{romero2014fitnets} (use shallower ones to initialize deeper ones) as well as shift covariance elimination using batch normalization~\cite{DBLP:conf/icml/IoffeS15}. The former is unnecessarily complex while the latter requires more GPU memory space since batch normalization could not be done in-place. Due to constrained resources, we instead apply \cite{he2015delving} that sets the standard deviation $std$ of the Gaussians used by weight initialization according to the formula $std=\sqrt{2/(k_l^2c_l)}$ in which $k_l$ and $c_l$ are the filter dimension and input channels of $l$-th convolutional layer.

To facilitate evaluation between architectures, we extract image features at the last fully connected layer (\texttt{fc7} layer for CaffeNet and FBNet, \texttt{loss1/fc} and \texttt{loss2/fc} and \texttt{pool5/7x7\_s1} for GoogleNet), do $L_2$ normalization, then train one-vs-rest linear SVM classifiers w.r.t classes of the datasets. Three last layers of GoogleNet are concatenated into a single vector. Parameter tuning for SVM training is done similarly to previous experiments. 

From the results presented in Table~\ref{table:deeper1k}, increasing the depth of a covnet greatly improves accuracy regardless of that convnet was trained from labeled data like  ImageNet or unlabeled data like FB-l. Also for the first time our FBNet outperforms CaffeNet trained on ImageNet. And FBNet is less performant than Vgg-16 which is deeper than ours. Our future work includes examining the performance of FBNet with even deeper structures which is expected to bridge the gap.

The results in Table~\ref{table:transfer1k} also reveals that GoogleNet is not so good as the conventional convnet architecture in learning transferable representation. This suggests us that a good form of transferability favors more degrees of distributed features which can be learned well using fully connected layers. However another reason may lie in the comparative feature dimensionality of features extracted from GoogleNet versus other convnets. We let this as a future work.
\begin{table}
\begin{centering}
\begin{tabular}{l|cc|cccc}
pretrained on & \multicolumn{2}{c}{IN-l} & \multicolumn{3}{c}{FB-l} \tabularnewline
\hline
architecture & CaffeNet & Vgg-16 & CaffeNet & FBNet & GoogleNet \tabularnewline
\hline
depth & 8 & 16 & 8 & 13 & 23 \tabularnewline
\hline\hline 
voc07 & 75.3 & 84.7 & 74.4 & 81.2 & 80.8 \tabularnewline
oxford102 & 87 & 87.5 & 87.6 & 88.8 & 86.8 \tabularnewline
mit67 & 60.8 & 70.0 & 58.4 & 64.7 & 59.6\tabularnewline
action40 & 60.2 & 72.6 & 55.9 & 63.3 & 63.8 \tabularnewline
caltech256 & 70.3 & 77.9 & 68.6 & 75.2 & 71.9 \tabularnewline
sun397 & 46.1 & 53.8 & 45.8 & 51 & 47.1 \tabularnewline
\end{tabular}
\par\end{centering}
\caption{Transferability of learned representations on 6 public datasets. In our approach (on the right), convnets are trained on the Web collection FB-l of 3,1 million images; the baselines (on the left) are trained on 1,2 million images of ImageNet ILSVRC challenge.}
\label{table:deeper1k}
\end{table}

\section{Application to Fine-grained Category Classification}\label{sec:finetune}
\noindent In this last section we examine the potential of using convnets pre-trained on Web images to specific problems. In particular three fine-grained category classification tasks are evaluated on convnets which are fine-tuned from the ones pre-trained with FB-l and IN-l. If both of them perform comparable then our approach is an efficient recipe in production. First collect a lot of Web images and train a convnet even the training data are noisy. This is followed by a subsequent refinement stage in which a smaller but clean dataset is used to fine-tune the previous convnet. This approach saves significantly annotation effort if successfully employed. 

Back to our experiment, three datasets are chosen: Birds~\cite{WahCUB_200_2011} (200 classes), Cars~\cite{krause20133d} (196 classes), and Flowers~\cite{Nilsback08} (102 classes). The numbers of training images may vary from 10 to 40 images per category.
\begin{figure}[!t]
\centering 
\subfloat[StanfordCars]{\includegraphics[width=\linewidth]{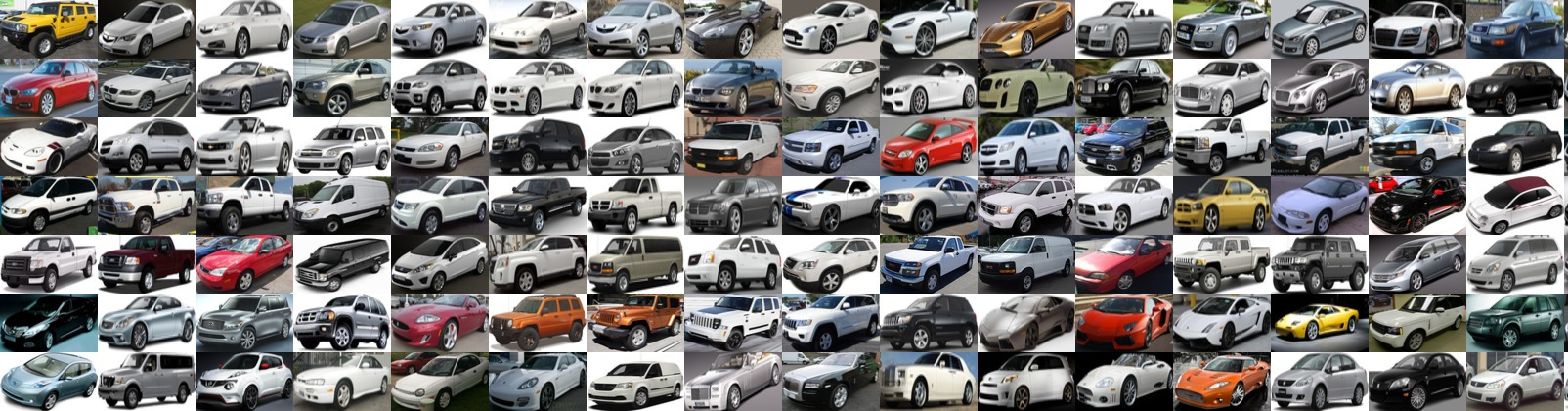}}\,
\subfloat[Oxford Flowers102]{\includegraphics[width=0.55\linewidth]{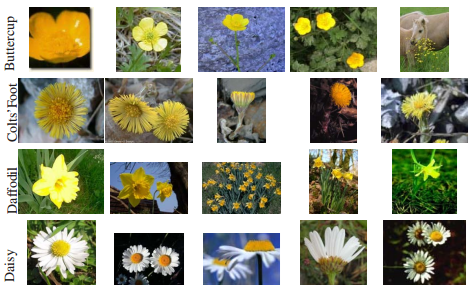}}\,
\subfloat[UCSD Bird200]{\includegraphics[width=0.4\linewidth]{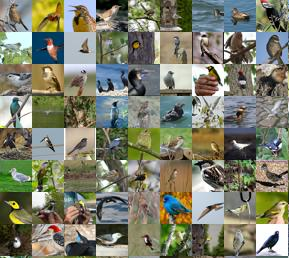}}\\
\caption{Fine-grained category datasets used in our experiments.}
\end{figure}

To measure the effectiveness of fine-tuning, classification accuracy is computed at two points before and after the process. The former is computed by evaluating linear SVM classifiers trained on normalized \texttt{fc7} features of a pre-trained net. The latter is straightforwardly computed based on output of the last layer of the tuned convnet. We adopt the CaffeNet architecture to for this experiment. The first 7 layers are initialized with pre-trained weights; the learning rate of each layer is $0.001$. The last fully connected layer is initialized with random weights; the learning rate is $0.01$. We use SGD with momentum $0.99$ and step size 2,000 iterations. Fine-tuning process is stopped after 10,000 iterations.
\begin{table}
\begin{centering}
\begin{tabular}{l|ll|ll}
pre-trained model & \multicolumn{2}{c}{FB-l} & \multicolumn{2}{c}{IN-l} \tabularnewline
\hline
fine-tuning & after & before & after & before \tabularnewline
 \hline\hline
birds~\cite{WahCUB_200_2011} & 52.7 & 47.1 & 52.1 & 47.1 \tabularnewline
cars~\cite{krause20133d} & 50.9 & 31.4 & 51.9 & 34.7 \tabularnewline
flowers~\cite{Nilsback08} & 86.7 & 87.6 & 86.4 & 87 \tabularnewline
\end{tabular}
\par\end{centering}
\caption{Classification accuracy of CaffeNet models trained on ImageNet (IN-l) and FlickrBing (FB-l). The accuracy before fine-tuning is computed by training linear SVM classifiers on \texttt{fc7} features extracted from corresponding models.}
\label{table:finecateg}
\end{table}

Results are shown in Table~\ref{table:finecateg}. In 3 out of 3 datasets, the accuracies of convnets tuned either by FB or Ref are quite comparable, before and after fine-tuning. In fact our net just performs slightly worse than the baseline in StanfordCars dataset and slightly outperforms in Flowers and Birds datasets. This is due to too small training size of Flower dataset (10 images per category) which is already explained in Section~\ref{sec:transfer1k}.

Based on observations above we speculate that the proposed approach proves itself as an economical and effective solution to supervised classification problems, especially in fine-grained category tasks. In order to improve further the performance, the fine-tuning weights should be initialized by a convnet trained from lots of Web images retrieved within a particular domain, for instance several thousands of flower categories. In this way the learned representation will be more discriminative to flora-related features. 

\section{Conclusion}\label{sec:conclu}
We proposed a novel approach that uses convnet to learn transferable image representation based on a  massive amount of (noisy) Web images. Throughout the paper we have successfully explored this approach under several problem scales, with different image reranking techniques, alternated several network architectures, and illustrated potential applications.

The significance of our study is threefold. First, our results show that convnets trained on Web images can obtain good generalization. Second, image reranking algorithms are useful to improve the generalization of convnet, especially at small and medium scales. In order to make image reranking useful at large scale problems, the training set should be considerably large and cover different visual variances of concepts. Third, deep convnet architectures can be trained on noisy images. Beside open problems addressed in this paper, we plan to investigate on how using a lot more unlabeled images can help reducing the need of labeled images in semi-supervised deep learning.
%

\section{References}
 \bibliographystyle{elsarticle-num}
 \bibliography{references}





\end{document}